\documentclass[letterpaper, 10 pt, conference]{ieeeconf}  %

\IEEEoverridecommandlockouts

\usepackage{cite}

\ifx\pdfoutput\undefined
\usepackage{graphicx}
\else
\usepackage[pdftex]{graphicx}
\fi

\usepackage{ifthen}
\usepackage[usenames,dvipsnames]{color}

\usepackage[font=small]{caption}

\usepackage{hyperref}

\usepackage{mathrsfs}     
\usepackage{amsmath}
\usepackage{amsfonts}
\usepackage{amssymb}
\usepackage{mathabx}
\usepackage{dsfont}
\usepackage{array}
\usepackage{wrapfig}
\usepackage{bigstrut}
\usepackage{multirow}
\usepackage{subcaption}
\usepackage{amssymb}
\usepackage{url}

\usepackage[ruled,linesnumbered,vlined]{algorithm2e}
\SetKwProg{Fn}{Function}{}{}
\SetArgSty{textrm}

\newtheorem{definition}{Definition}

\SetAlFnt{\small}

\begin{document}
%
\title{\LARGE \bf Footstep and Motion Planning in Semi-unstructured Environments Using Randomized Possibility Graphs}


\author{Michael X. Grey$^{1}$ \and Aaron D. Ames$^{2}$ \and C. Karen Liu$^{3}$
\thanks{$^{1}$Michael X. Grey is a doctoral candidate in the School of Interactive Computing,
            Georgia Institute of Technology, Atlanta, GA 30332
            {\tt\small mxgrey@gatech.edu}}
\thanks{$^{2}$Aaron D. Ames is a Bren Professor in Mechanical and Civil Engineering \& Control and Dynamical Systems,
            California Institute of Technology, Pasadena, CA 91125
            {\tt\small ames@caltech.edu}}
\thanks{$^{3}$C. Karen Liu is an Associate Professor in the School of Interactive Computing,
            Georgia Institute of Technology, Atlanta, GA 30332
            {\tt\small karenliu@cc.gatech.edu}}
}

\maketitle

\begin{abstract}

Traversing environments with arbitrary obstacles poses significant challenges for bipedal robots. In some cases, whole body motions may be necessary to maneuver around an obstacle, but most existing footstep planners can only select from a discrete set of predetermined footstep actions; they are unable to utilize the continuum of whole body motion that is truly available to the robot platform. Existing motion planners that can utilize whole body motion tend to struggle with the complexity of large-scale problems. We introduce a planning method, called the ``Randomized Possibility Graph'', which uses high-level approximations of constraint manifolds to rapidly explore the ``possibility'' of actions, thereby allowing lower-level motion planners to be utilized more efficiently. We demonstrate simulations of the method working in a variety of semi-unstructured environments. In this context, ``semi-unstructured'' means the walkable terrain is flat and even, but there are arbitrary 3D obstacles throughout the environment which may need to be stepped over or maneuvered around using whole body motions.

\end{abstract}

\IEEEpeerreviewmaketitle

\section{Introduction}

\begin{figure}
  \begin{subfigure}[b]{0.48\linewidth}
    \captionsetup{justification=justified}
    \centering
    \includegraphics[width=\linewidth]{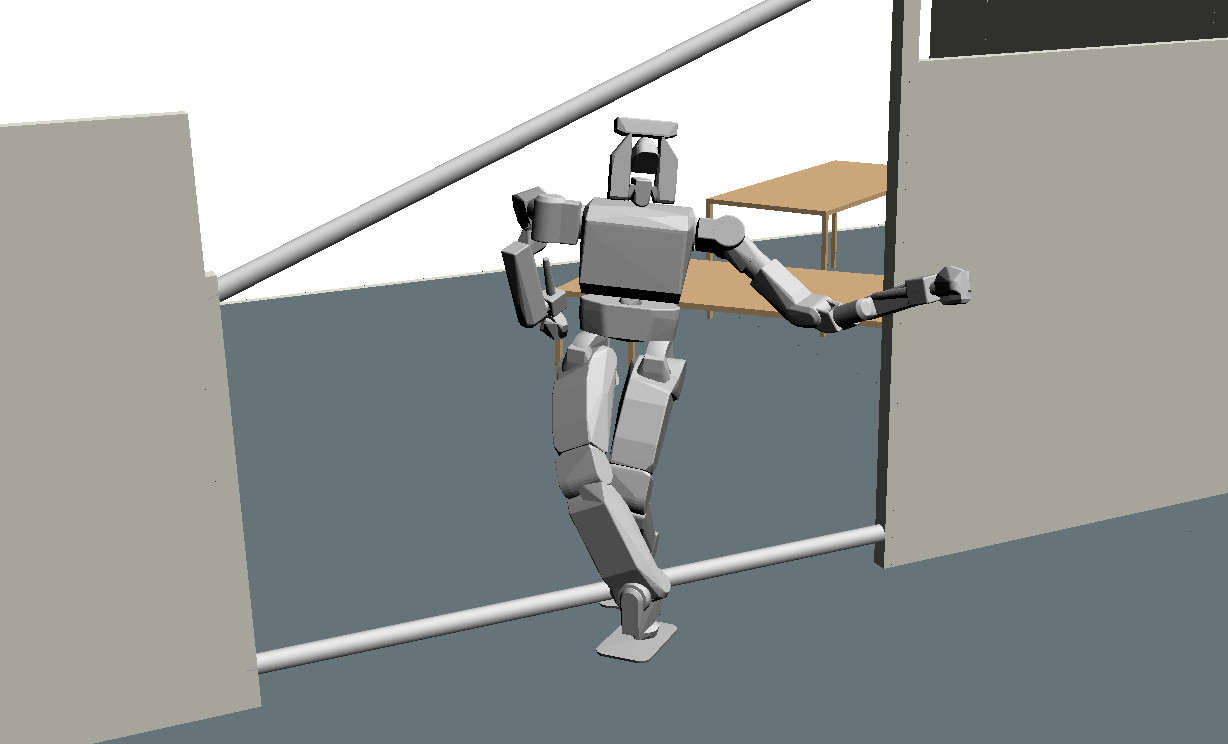}
    \caption{North Doorway: Fit between the high and low bars}
    \label{fig:north_door}
  \end{subfigure}
  \begin{subfigure}[b]{0.48\linewidth}
    \captionsetup{justification=justified}
    \centering
    \includegraphics[width=\linewidth]{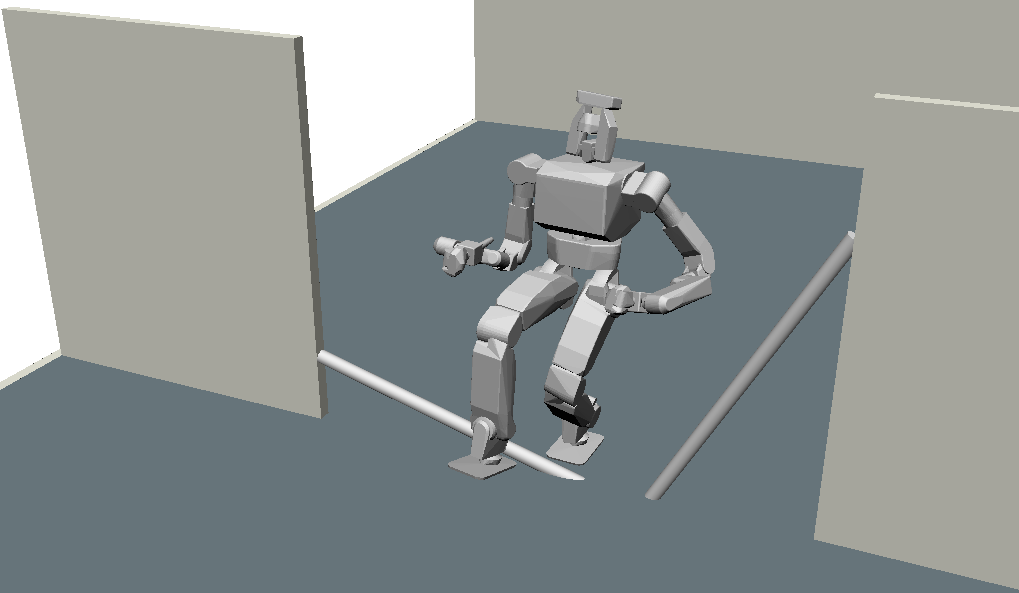}
    \caption{South Doorway: Step over the arbitrarily angled bars}
    \label{fig:south_door}
  \end{subfigure}
  
  \begin{subfigure}[b]{0.48\linewidth}
    \captionsetup{justification=justified}
    \centering
    \includegraphics[width=\linewidth]{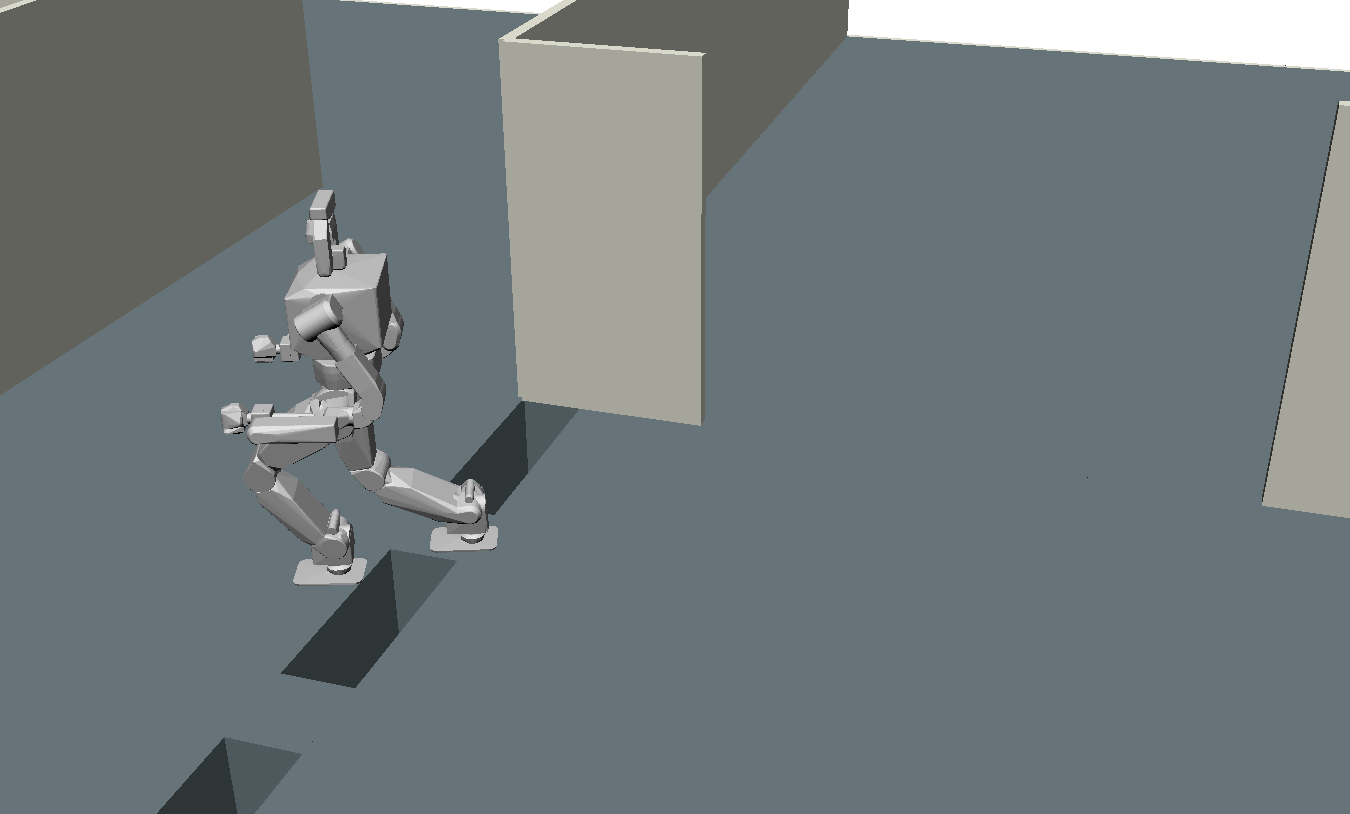}
    \caption{West Doorway: Step across the gaps without falling in}
    \label{fig:west_door}
  \end{subfigure}
  \begin{subfigure}[b]{0.48\linewidth}
    \captionsetup{justification=justified}
    \centering
    \includegraphics[width=\linewidth]{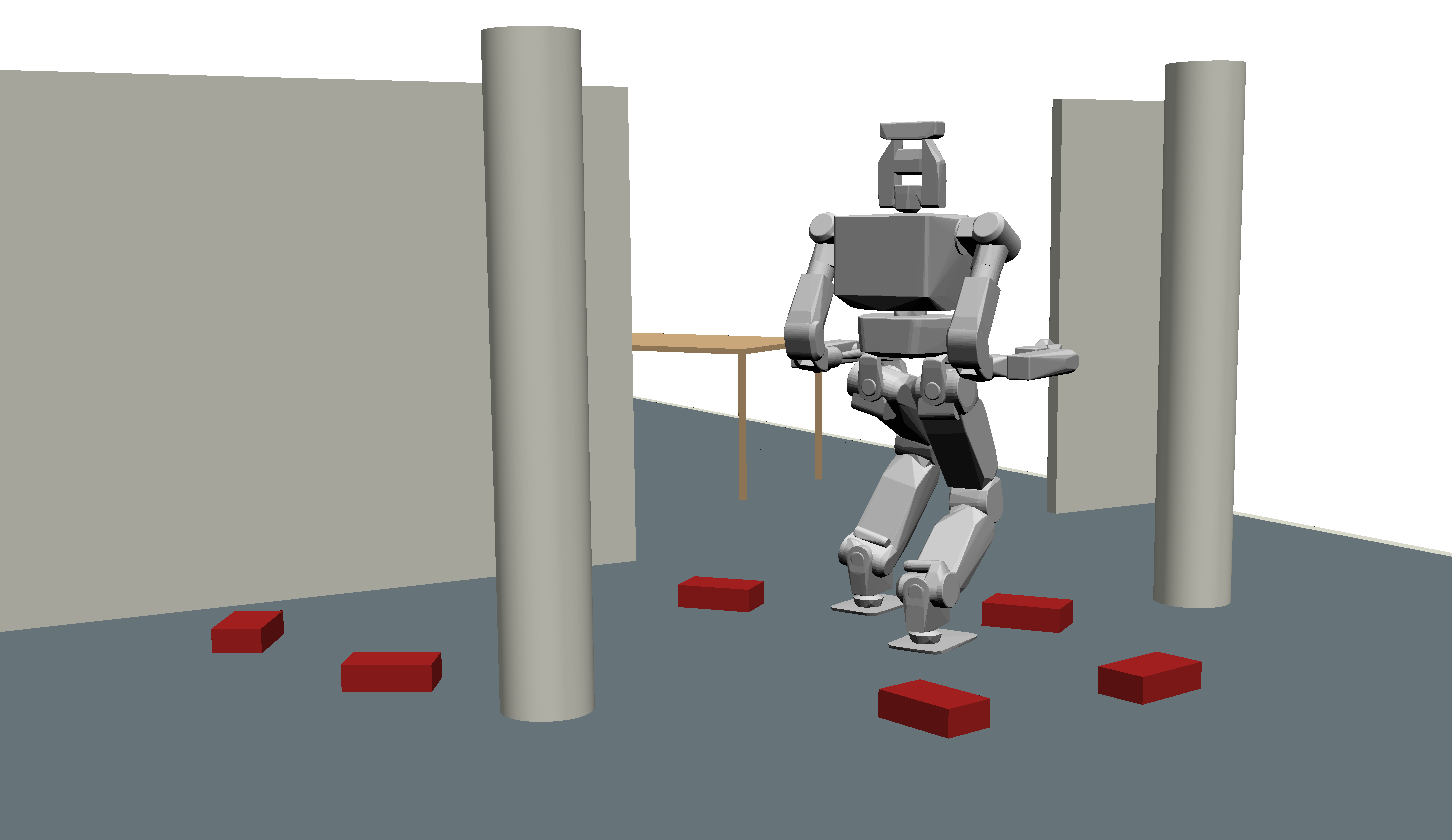}
    \caption{East Doorway: Step over the arbitrarily laid out bricks}
    \label{fig:east_door}
  \end{subfigure}

  \caption{Doorways from the ``Four Routes Scenario''. Each passage has unique challenges that the robot must overcome.}
  \label{fig:doorways}
\end{figure}

As humanoid robotics technology continues to advance, there is growing interest in using legged platforms for disaster relief in hazardous environments where wheeled platforms may be unable to traverse. These hazardous environments may include structures like nuclear reactors or collapsing buildings. We refer to these environments as ``semi-unstructured'', because they were originally designed to be easily navigable but may have succumbed to conditions where intended routes are no longer clear of obstructions. Perhaps the passages have been structurally damaged or are being littered by various obstacles that have been left behind. For a legged robot to be useful under these conditions, it may need to use strategic foot placement and utilize the entire range of motion of its body, such as maneuvering between the bars in Fig. \ref{fig:north_door}.

Footstep planning is a cornerstone of humanoid robotics research due to its importance in navigating challenging terrain. Many algorithms have been designed for planning footsteps, but most tend to fall into one of four categories:

\begin{enumerate}
\item Planners that use bounding boxes for navigation
\item Planners that choose from a discrete set of actions
\item Planners that use optimization methods
\item Planners that find paths through and between discretized sets of contacts
\end{enumerate}

Each of these categories has a limited scope of applicability. An example of the first category is \cite{pettre20032} where the lower body is given a bounding box that encapsulates all the potential leg motion that the humanoid agent might exhibit. The upper body is left unbounded with the assumption that any upper body collisions will be resolved in a second stage. As a result, the humanoid is unable to realize when it could step over a small obstacle. By contrast, in \cite{perrin2012real}, each foot is given an independent bounding box, and then the remainder of the robot's geometry is given one large bounding box which is elevated off the ground. All of the robot's motion takes place within these three bounding boxes, but having the bulk of the bounding geometry elevated allows the robot to navigate over small obstacles. However, the robot is unable to maneuver its upper body around overhanging obstacles.

Examples of work belonging to the second category include \cite{kuffner2001footstep, kuffner2003online, chestnutt2003planning, garimort2011humanoid, perrin2012fast}. In all of these methods, there is a discrete set of predetermined actions that are available to the planner. They often use A* search where the predetermined actions are used for branching. These methods tend to be effective at stepping over and around short obstacles or gaps, but they generally are not able to alter their upper body configurations to negotiate overhanging obstacles or tight passages.

A variety of optimization methods have been applied to the problem of footstep planning and motion generation. In \cite{fallon2015architecture} the footstep planning component was done with a human-in-the-loop, and then a motion trajectory was optimized over those footstep locations. The human operator is taken out of the loop in \cite{deits2014footstep} and proofs for optimality and completeness are available, but this comes at the cost of requiring convex collision-free regions for the robot to pass through. Optimization can also be used to generate dynamically stable periodic walking gaits and controllers, such as methods proposed in \cite{hereid20163d, powell2012motion}, although these do not address global search problems.

The fourth category is the only one where probabilistically complete whole body motion planning is available. For this category, a mode is defined by a set of contacts between the robot and the environment. These contacts determine the feasibility constraints for any configurations associated with that mode. Methods in this category are usually used to navigate extremely rough terrain or plan whole body manipulation. As an example, Multi-modal PRM \cite{hauser2009multi}, Random-MMP \cite{hauser2011randomized}, and related methods \cite{hauser2008motion, bretl2004multi} first sample modes in the environment and then generate whole body motions to transition through and between those modes. A configuration that transitions between two modes must satisfy the feasibility constraints of both modes simultaneously. Random-MMP is theoretically capable of navigating the semi-unstructured environments of this paper, but it struggles to find narrow passages, which may be important for navigating through challenging environments.


In this paper, we present the ``Randomized Possibility Graph'' (RPG) for efficient whole body motion planning in semi-unstructured environments. By exploring the \emph{possibility} of actions within a continuous search space, we can rapidly expand the search without being bottlenecked by expensive motion planning queries. We find that the RPG is effective at identifying passages which would be considered narrow by Random-MMP. We show that information from the RPG can be used to guide Random-MMP through narrow passages by focusing its sampling behavior. Together, these algorithms are able to traverse semi-unstructured environments on the order of seconds to minutes, depending on the complexity of the environment.

\section{Randomized Possibility Graph}\label{sec:posgraph}

Motion planning methods ordinarily operate by constructing graphs or trees which consist of configurations that fully exist within the feasibility constraint manifold of the action they are performing. Remaining within this manifold is a reasonable requirement to place on the graph, because any vertices or edges which lie outside of the manifold are, by definition, invalid---which may mean it is physically impossible, or simply harmful to the robot or its surroundings. Unfortunately, for a humanoid robot to remain on the constraint manifold, expensive calls to a whole body inverse kinematics (IK) solver must be performed \cite{sentis2006whole, sugihara2002whole, gienger2005task}. This results in a critical bottleneck if a broad area needs to be explored before finding a solution. 

The key idea of this work is to explore the \emph{possibility} of an action first, instead of immediately committing to costly whole body inverse kinematics queries. Therefore, the problem is broken down into two stages: The exploration of possibilities $\mathcal{P}$, and the generation of motions $\mathcal{M}$. The high-level graph generated during $\mathcal{P}$ will guide the efforts of the low-level planners used by $\mathcal{M}$.

The governing logical principles behind the RPG have a theoretical grounding in Possibility Theory \cite{dubois2012possibility}, but the concepts are intuitive enough that a knowledge of Possibility Theory is not necessary to proceed. It is enough to understand that the \emph{possibility} of any given action $e$ can be labelled with ``impossible'', ``possible'', or ``indeterminate'' depending on whether it satisfies the necessary conditions ($C_N$) or the sufficient conditions ($C_S$) that are assigned to it:
\begin{equation}
e.\text{label} = 
\begin{cases}
    \text{``impossible''} & \quad \text{if } C_N(e) \text{ is false} \\
    \text{``possible''} & \quad \text{if } C_S(e) \text{ is true} \\
    \text{``indeterminate''} & \quad \text{otherwise } \\
  \end{cases}
\end{equation}

If we design necessary and sufficient conditions that can be checked much more quickly than querying the original constraint manifold, we can then construct a \emph{Randomized Possibility Graph}, whose vertices are connected by either ``possible''  or ``indeterminate'' edges, and expand it very efficiently for whole body motion planning.

\subsection{Simplifying the Manifold: Sufficient vs. Necessary}

\begin{figure}
  \centering
  \includegraphics[width=\linewidth]{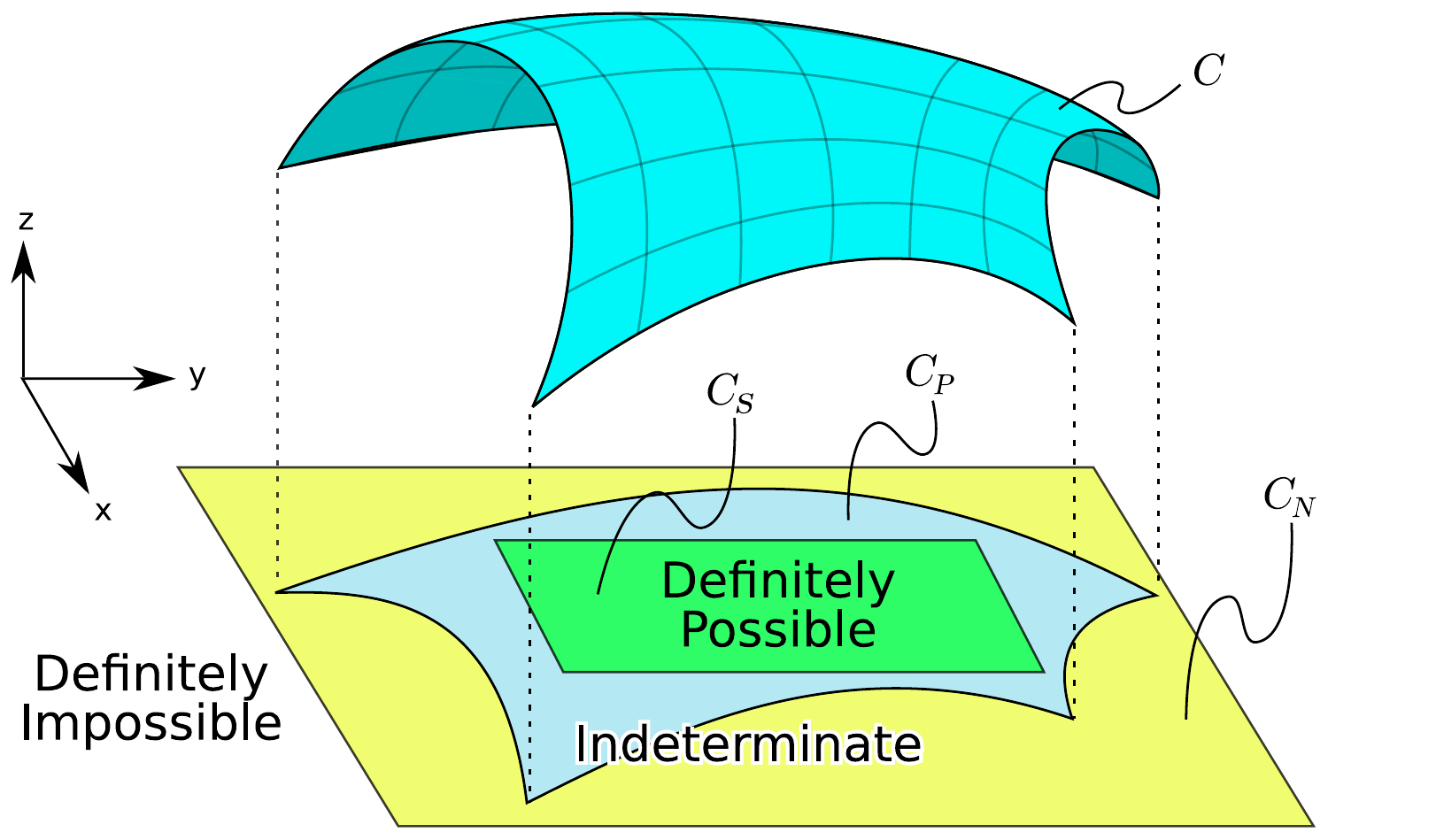}
  
  \caption{\label{fig:abstract_manifold}Visual depiction of an abstract constraint manifold, $C$ and its projection. The manifold is projected, $C_P$, from 3D space onto a plane. ``Sufficient'' $C_S$ and ``Necessary'' $C_N$ boundaries are fitted within and around the projection of the manifold. Elements inside the green box are definitely ``Possible''. Elements outside of the yellow box are definitely ``Impossible''. Elements inside the yellow box but outside of the green box are ``Indeterminate'' because they might or might not lie on the projection. Identifying whether a point lies inside or outside of $C_S$ or $C_N$ may be considerably faster than identifying whether it lies inside or outside of $C_P$.}
\end{figure}

To construct the RPG, we must first design sufficient and necessary conditions for the feasibility constraint manifold of the action whose possibilities we are exploring. We should design the conditions to be tested quickly in order to reduce the computational cost of exploration. The conditions should also use as few parameters as possible, because randomized sampling methods are more effective in low-dimensional search spaces.

Suppose we have a 2D constraint manifold, $C$, which exists in a 3D state space (Fig. \ref{fig:abstract_manifold}). Let the $xy$-plane be a low-dimensional feature space which contains the essential information for navigating $C$. We denote the projection of $C$ by $C_P$. Even with a flattened-out projection, identifying which points are inside or outside of the manifold may still be costly or difficult, because the boundary of $C_P$ may consist of functions that are expensive to compute or hard to fully define. However, suppose a box, circle, or some other simple shape can be fit within $C_P$ such that it is \emph{guaranteed} that every point within the simple shape also lies within the manifold projection. Such a shape would be a suitable representation of the sufficient condition manifold, $C_S$. Any point lying inside of $C_S$ also lies inside of $C_P$ and should be labelled with ``possible''. Similarly, if $C_P$ can be bounded by a simple shape, $C_N$, such that $C_P \subseteq C_N$, then $C_N$ would qualify as the necessary condition manifold. Any point lying outside of $C_N$ should be labelled with ``impossible''. Finally,  any point inside of $C_N$ but outside of $C_S$ should be labelled with ``indeterminate''.

\subsection{RPG Construction}

In the previous section, we introduced the concepts of $C_N$ and $C_S$, the necessary and sufficient (respectively) condition manifolds which occupy a lower dimensional space than the state space. We define $\mathscr{E} = P(X)$ to be the ``possibility exploration space'' where $X$ is the state space of the robot, and in this paper $P : X \to \text{SE(3)}$ maps from the robot's state to the SE(3) transformation of the robot's root link. In general, $P$ is chosen to be a projection operator such that $C_S \subseteq C_N \subseteq P(X)$. We will use $\mathscr{E}$ as the search domain for the possibility exploration stage $\mathcal{P}$.

\begin{definition}
\normalfont
\label{def:posgraph}A Randomized Possibility Graph is a tuple
\begin{displaymath}
RPG = (\Gamma_\mathcal{P}, \Phi_\mathcal{M}, \Omega_\mathcal{M}, \Gamma_\mathcal{M})
\end{displaymath}
where,
\begin{itemize}
\item $\Gamma_\mathcal{P} = (V_\mathcal{P}, E_\mathcal{P})$ is a graph where $V_\mathcal{P}$ is a set of vertices which are elements of $\mathscr{E}$, and $E_\mathcal{P}$ is a set of directed edges, each with a ``possible'' or ``indeterminate'' label,
\item $\Phi_\mathcal{M} : E_\mathcal{P} \times E_\mathcal{P} \to X^k$ is an operator which takes two edges and produces a set of $k > 0$ states,
\item $\Omega_\mathcal{M} : X \times E_\mathcal{P} \times X \to X^f$ is an operator which maps two states with a possibility edge in between them into a discretized trajectory of $f \geq 0$ states, where $f = 0$ implies failure to find a trajectory,
\item $\Gamma_\mathcal{M} = (V_\mathcal{M}, E_\mathcal{M})$ is a graph where $V_\mathcal{M}$ is a set of vertices which are elements of $X$, and $E_\mathcal{M}$ is a set of directed edges which indicate feasible paths between the vertices of $V_\mathcal{M}$.
\end{itemize}
\end{definition}

\begin{figure}
  \centering
  \begin{subfigure}[b]{\linewidth}
    \captionsetup{justification=justified}
    \centering
    \includegraphics[width=0.76\linewidth]{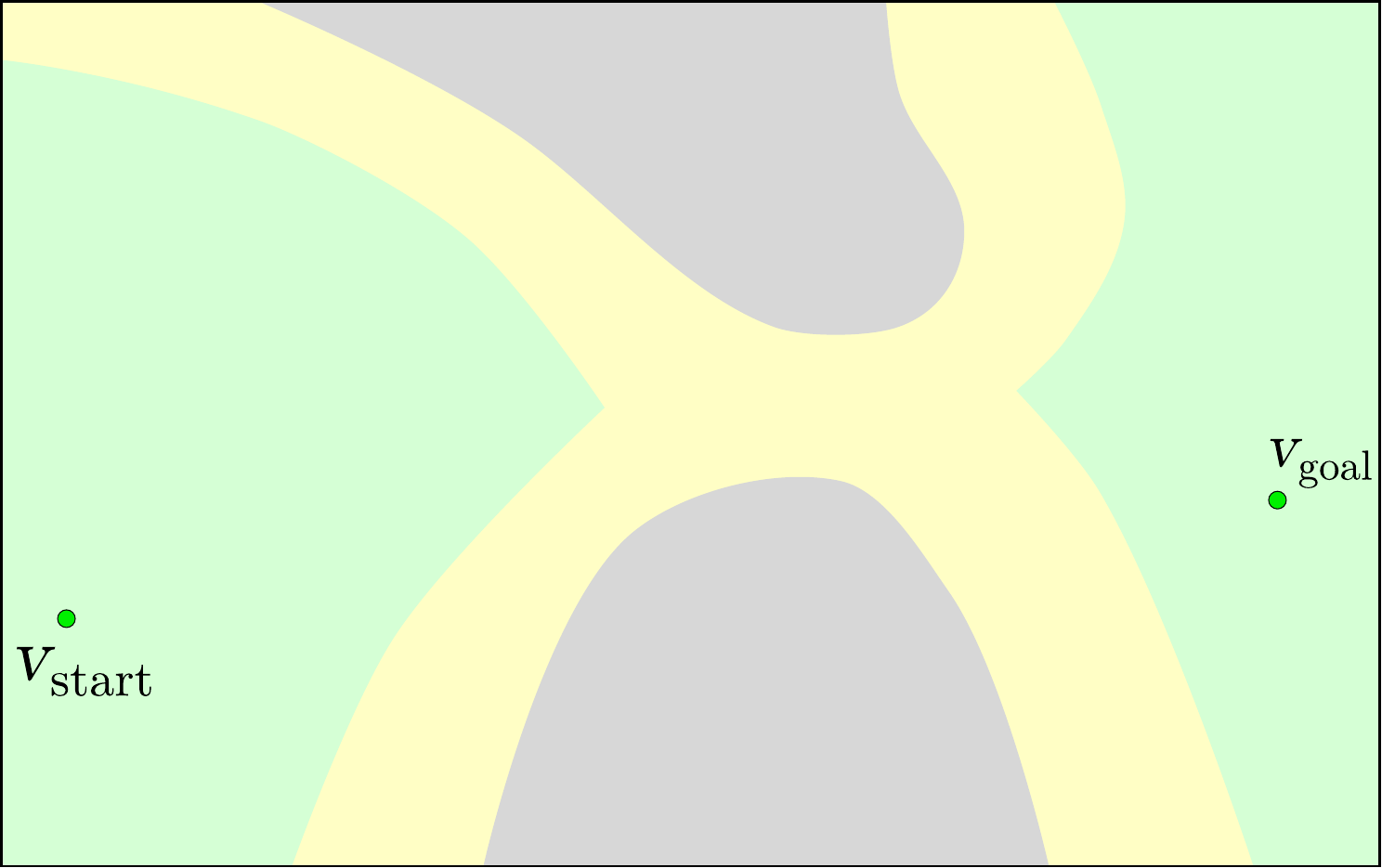}
    \caption{The RPG is initialized with just a start vertex and goal vertex. The pale green regions are areas where $C_S$ is satisfied, pale yellow is where $C_N$ is satisfied, and gray is where $C_N$ is violated. These colored regions represent information which is not directly known to the algorithm, so it must be searched via randomized sampling.}
    \label{fig:rpg_process_0}
  \end{subfigure}
  \begin{subfigure}[b]{\linewidth}
    \captionsetup{justification=justified}
    \centering
    \includegraphics[width=0.76\linewidth]{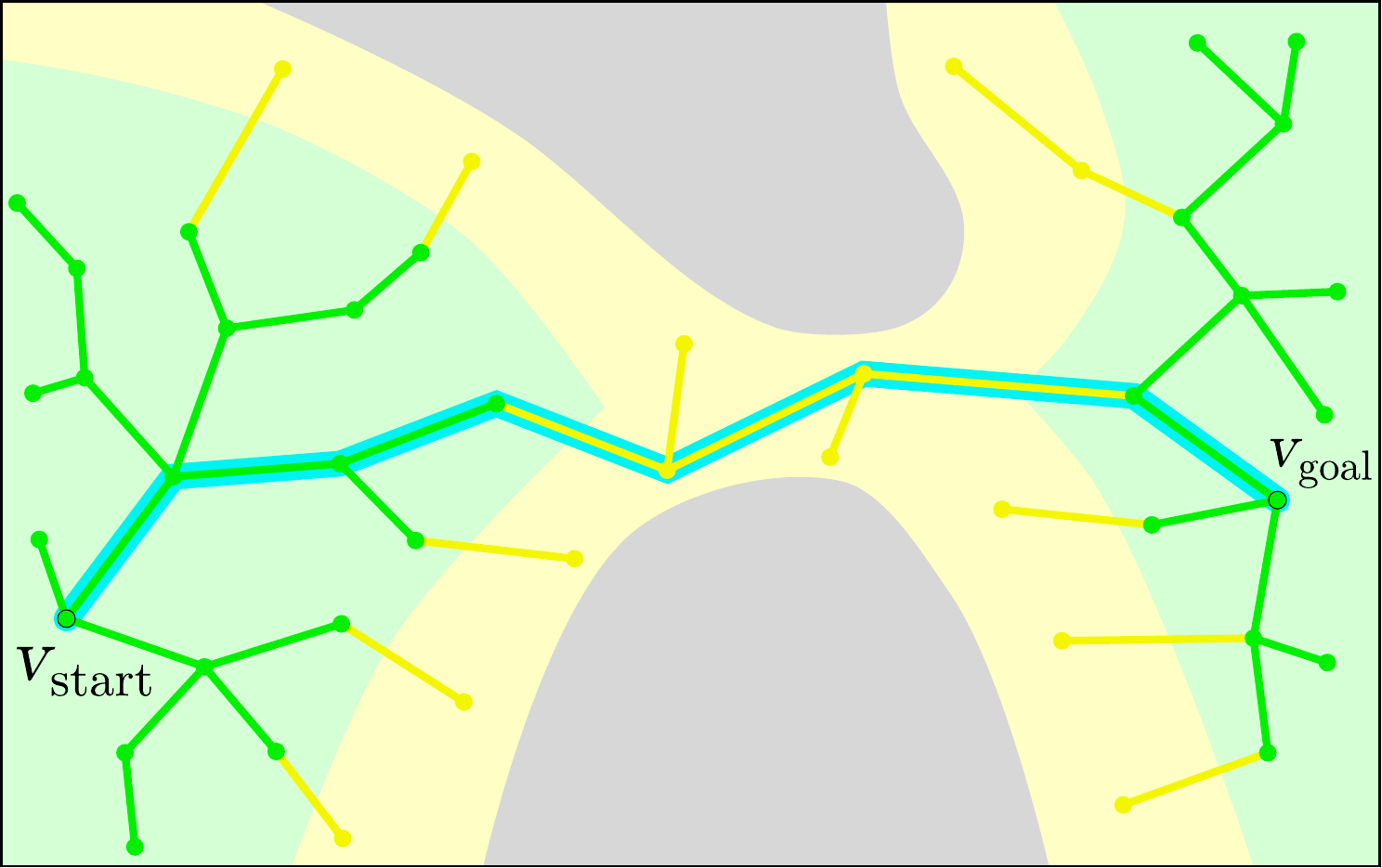}
    \caption{A randomized sampling motion planner has constructed $\Gamma_\mathcal{P}$ to find a path from the start to the goal through $\mathscr{E}$. This provides a guide route, which is highlighted in cyan.}
    \label{fig:rpg_process_1}
  \end{subfigure}
  \begin{subfigure}[b]{\linewidth}
    \captionsetup{justification=justified}
    \centering
    \includegraphics[width=0.76\linewidth]{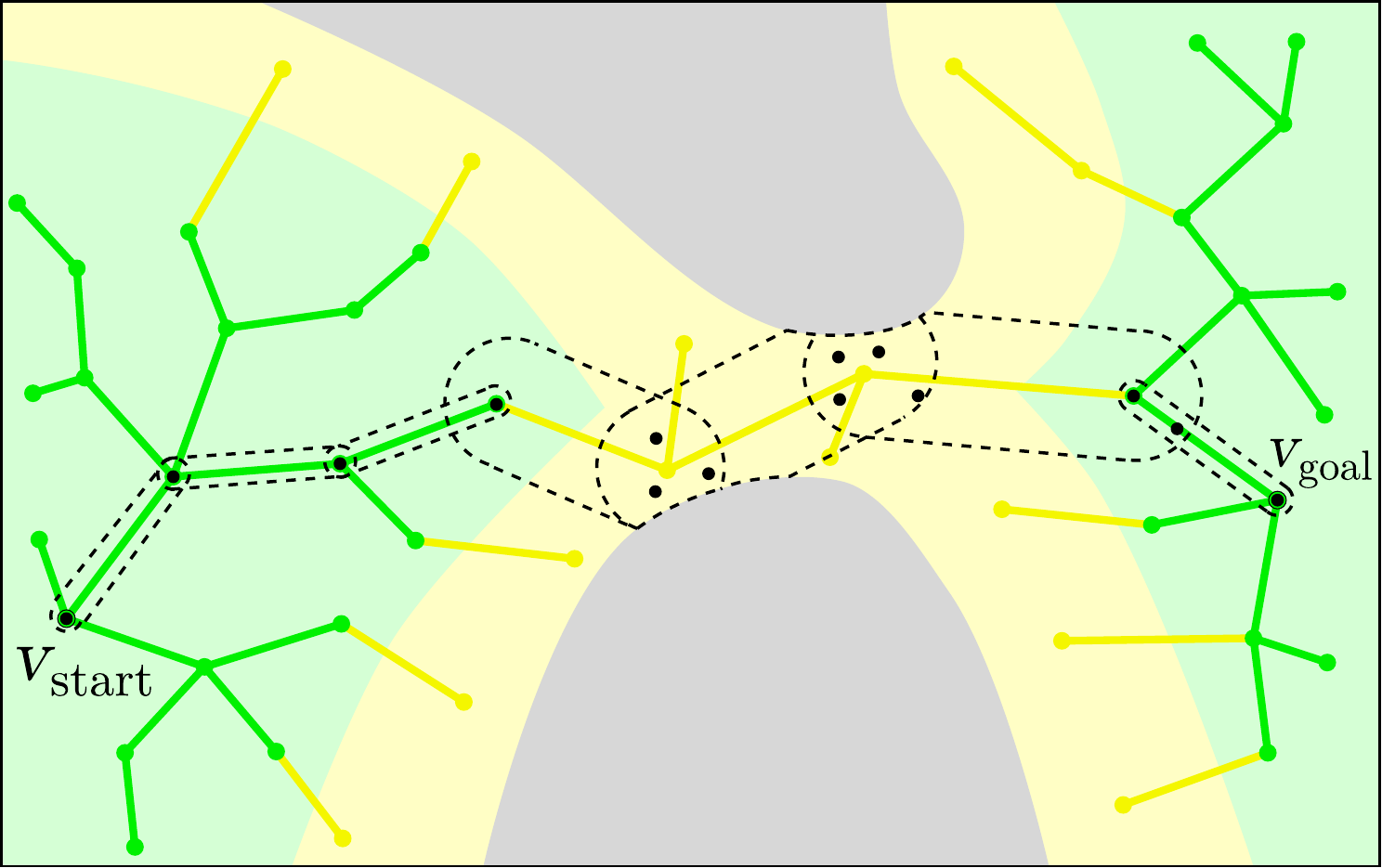}
    \caption{The dotted regions represent the ranges of $\pi_S$ or $\pi_N$ for each edge along the guide route. The $\pi_S$ regions are thin because we are guaranteed to find a solution directly along the routes where $C_S$ is satisfied. Conversely, $\pi_N$ may need to search a broader area to find a solution. $\Phi_\mathcal{M}$ is applied to the edges of the guide route, generating sets of states (black dots) where the ranges of the planners overlap.}
    \label{fig:rpg_process_2}
  \end{subfigure}
  \begin{subfigure}[b]{\linewidth}
    \captionsetup{justification=justified}
    \centering
    \includegraphics[width=0.76\linewidth]{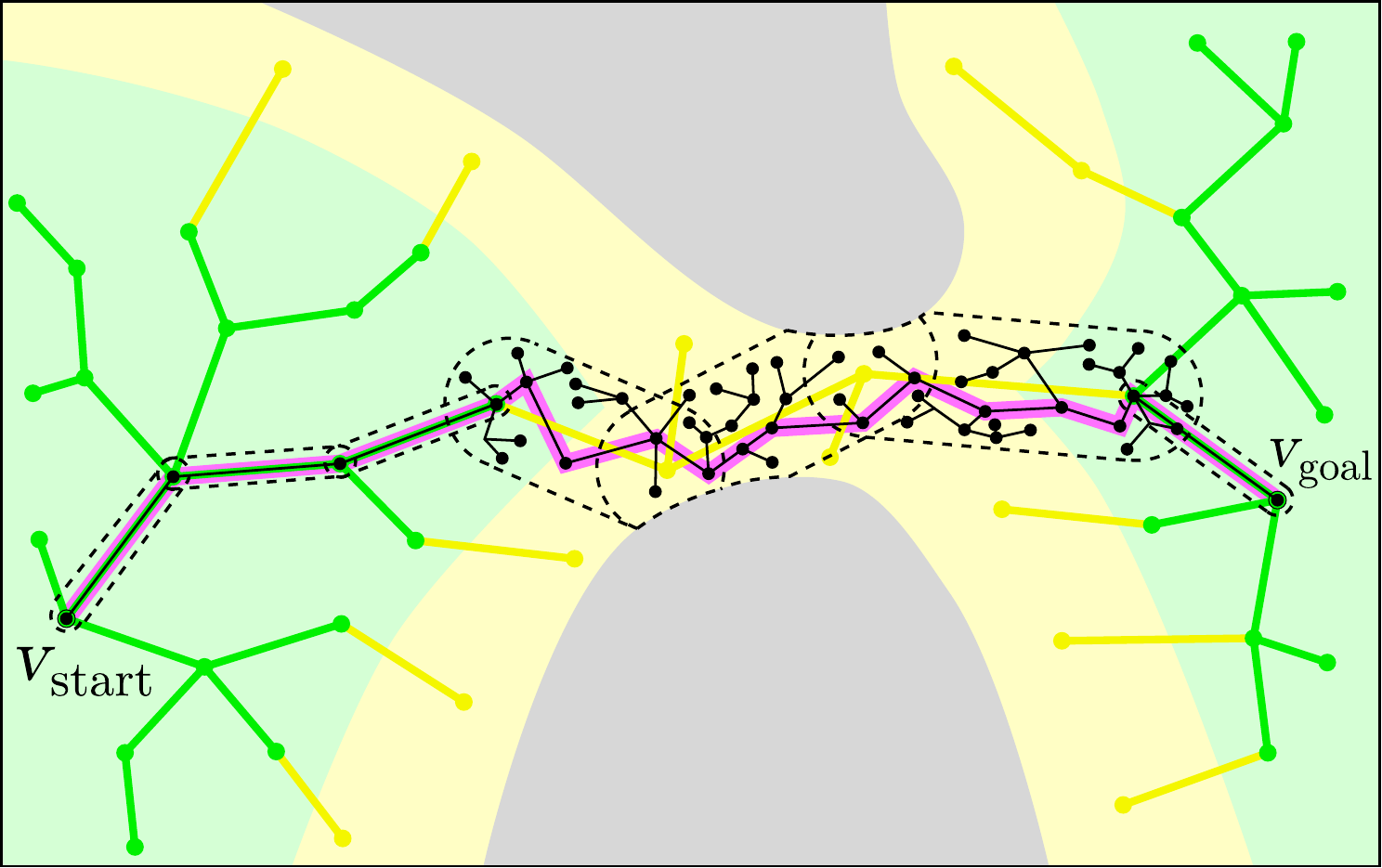}
    \caption{States and whole body paths (black dots and edges) are generated by $\Omega_\mathcal{M}$ using the guide route and the states generated by $\Phi_\mathcal{M}$. These elements are put into $\Gamma_\mathcal{M}$ until a complete whole body path is found from the start to the goal.}
    \label{fig:rpg_process_3}
  \end{subfigure}
  
  \caption{Illustration of the RPG procedure}
  \label{fig:rpg_process}
\end{figure}

The graph $\Gamma_\mathcal{P}$ is used to solve the first stage, $\mathcal{P}$, which explores possibilities. $\Gamma_\mathcal{P}$ can be constructed using a sample-based motion planner, such as PRM \cite{kavraki1996probabilistic} or RRT \cite{kuffner2000rrt}. If $C_N$ or $C_S$ have a small volume within $\mathscr{E}$, then a projection-based sampler may be needed, such as CBiRRT \cite{cbirrt} where $C_N$ and $C_S$ are treated as task constraints. Alternatively, $C_N$ and $C_S$ can be treated as ``hard'' and ``soft'' task constraints respectively (meaning that $C_N$ is required but $C_S$ is preferred) which would make the method in \cite{kunz2012manipulation} more applicable. For this paper, we use the ``hard'' and ``soft'' constraint approach, and we refer to this high-level planner as $\Pi_\mathcal{P}$.

We initiate $\Gamma_\mathcal{M}$ with one or more ``start'' states, $V_\mathcal{M}^\text{start}$, and one or more ``goal'' states, $V_\mathcal{M}^\text{goal}$. $\Gamma_\mathcal{P}$ is initiated with the projections of these states, $P(V_\mathcal{M}^\text{start})$ and $P(V_\mathcal{M}^\text{goal})$. The high-level motion planner of choice, $\Pi_\mathcal{P}$, is used to find a route $(e_1,...,e_n \in E_\mathcal{P})$ which connects a start projection to a goal projection while remaining within $C_N$. The solved path generated by $\Pi_\mathcal{P}$ represents a guide route, similar to the ``guide trajectories'' of \cite{tonneau2015reachability}. An illustration of this process can be seen in Figs. \ref{fig:rpg_process_0}-\ref{fig:rpg_process_1}. The guide route is used to focus the efforts of the next stage, $\mathcal{M}$.


Provided the guide route found by $\mathcal{P}$, we want to use $\Omega_\mathcal{M}$ to examine each segment of the route to see if it can be realized in $X$. We define $\Omega_\mathcal{M}$ as:
\begin{displaymath}
\Omega_\mathcal{M}(x_i^0, e_i, x_i^f) =
\begin{cases}
    \pi_S(x_i^0, e_i, x_i^f) & \quad \text{if } C_S(e_i) \\
    \pi_N(x_i^0, e_i, x_i^f) & \quad \text{if } C_N(e_i) \land \neg C_S(e_i)
  \end{cases}
\end{displaymath}
where $\pi_S$ and $\pi_N$ are sub-planners which produce full state space trajectories given a sub-start state ($x_i^0$), a sub-goal state ($x_i^f$), and an edge ($e_i$) of $E_\mathcal{P}$ which is used as a ``guide''. The ``guide'' edge may be used to compute a heuristic, determine footstep locations, or focus randomized samples, depending on the nature of the planner. We choose $\pi_S$ such that it is guaranteed to quickly find a solution along routes where the sufficient conditions $C_S$ are satisfied. $\pi_N$ is a whole body motion planner, ideally with a probabilistic completeness guarantee. $\pi_N$ can be applied to edges which satisfy the necessary conditions $C_N$ even if the sufficient conditions are not satisfied, but it is not guaranteed to return a solution.

To find a feasible continuous motion through state space, we need to choose $x_i^f$ and $x_{i+1}^0$ to be equal. Define $X_0(e_i)$ to be the set of states that can be used as sub-start states, $x_i^0$, for $\Omega_\mathcal{M}(x_i^0, e_i, x_i^f)$, and $X_f(e_i)$ to be the set that can be used as sub-goal states, $x_i^f$. We can then define $\Phi_\mathcal{M}(e_i, e_{i+1}) = X_f(e_i) \cap X_0(e_{i+1})$. Using $\Phi_\mathcal{M}$ to determine the endpoints (sub-starts and sub-goals) used by $\Omega_\mathcal{M}$ ensures that the RPG is able to create continuous state trajectories. Note that $\Phi_\mathcal{M}(e_0, e_1) = V_\mathcal{M}^\text{start} \cap X_0(e_1)$, and $\Phi_\mathcal{M}(e_n, e_{n+1}) = X_f(e_n) \cap V_\mathcal{M}^\text{goal}$. $\Phi_\mathcal{M}$ is illustrated by the black dots in the overlapping dotted regions of Fig. \ref{fig:rpg_process_2}. Determining $X_0(e)$ and $X_f(e)$ will depend on the implementation of $\pi_S$ and $\pi_N$, but most planners have either a discrete set of permissible endpoints or a continuous set that can be sampled from. In practice, many planners allow multiple start and goal states to be specified per query.


The overall procedure for planning with the RPG is to generate a guide route $(e_1, ..., e_n \in E_\mathcal{P})$ by constructing $\Gamma_\mathcal{P}$ with $\Pi_\mathcal{P}$ and feeding that route through $\Phi_\mathcal{M}$ and then through $\Omega_\mathcal{M}$. Whenever $\Omega_\mathcal{M}$ identifies feasible state trajectories, the states and edges of those trajectories are added to $\Gamma_\mathcal{M}$. A solution is found when $\Gamma_\mathcal{M}$ contains a path from a start to a goal state. The procedure is illustrated in Fig. \ref{fig:rpg_process}.

Depending on the implementation of $\pi_N$, it may take an indeterminable amount of time to produce a solution. Moreover, it might not be able to produce a solution for some $e_i$ if the edge is not a truly feasible guide route. Rather than waiting for $\Omega_\mathcal{M}$ to return a result of success or failure, the stage $\mathcal{P}$ can search for alternative guide routes by deleting any of the indeterminate edges of the guide route from $\Gamma_\mathcal{P}$ and then continuing to grow $\Gamma_\mathcal{P}$ in parallel to $\mathcal{M}$. This parallelism allows the RPG to avoid being bottlenecked by challenging routes when alternatives exist. When elements are added to $\Gamma_\mathcal{M}$, their projections can be added to $\Gamma_\mathcal{P}$ as ``possible`` elements to assist the ongoing high-level search.

\section{Walking in Semi-unstructured Environments}\label{sec:gmmp}

\begin{figure}[h]
  \centering
  \begin{subfigure}[b]{0.47\linewidth}
    \captionsetup{justification=justified}
    \centering
    \includegraphics[width=0.94\linewidth]{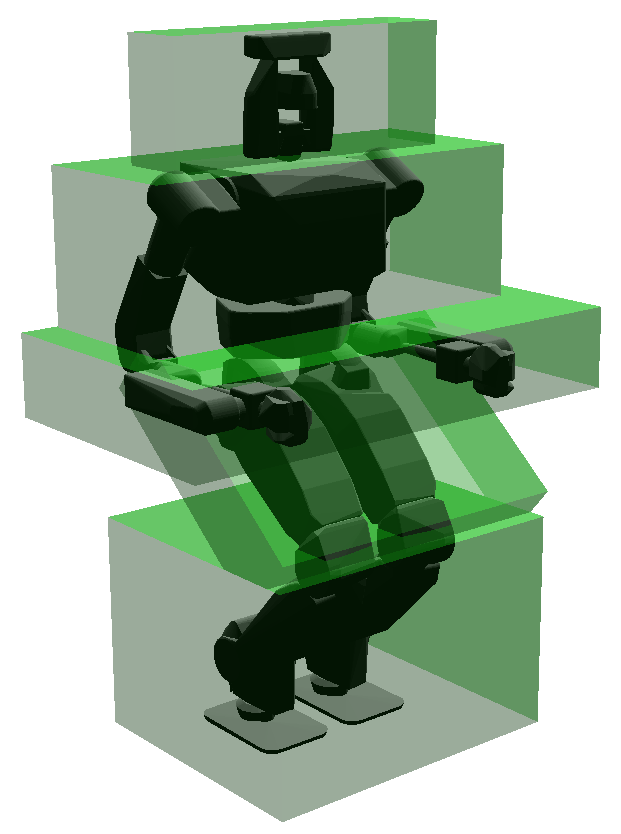}
    \caption{Bounding geometry used for sufficient conditions (Green)}
    \label{fig:bounding_geometry}
  \end{subfigure}
  \hfill
  \begin{subfigure}[b]{0.47\linewidth}
    \captionsetup{justification=justified}
    \centering
    \includegraphics[width=0.94\linewidth]{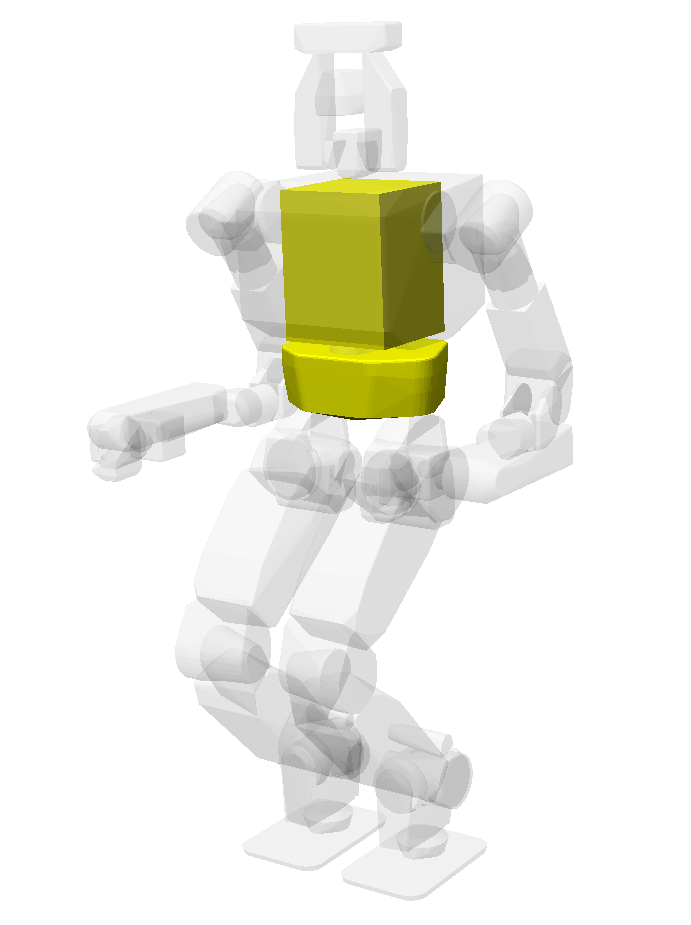}
    \caption{Minimal geometry used for necessary conditions (Yellow)}
    \label{fig:minimal_geometry}
  \end{subfigure}
  
  \caption{Collision geometries used for the (a) sufficient and (b) necessary conditions. Since (a) occupies more space, the constraint manifold for the sufficient conditions is smaller---and therefore more restrictive---than the manifold for the necessary conditions.}
  \label{fig:sweep}
\end{figure}


We seek to find feasible motion plans for a bipedal robot to traverse a ``semi-unstructured'' environment. In the context of this paper, we define ``semi-unstructured'' to mean that the terrain is structured---flat and even---but there are arbitrary unstructured obstacles throughout the environment. The robot may need to step over, duck under, or maneuver around these obstacles using whole body motions. The planner is provided with a kinematic model of the robot, a geometric model of all obstacles---including walls---and a layout of the floor. No other contextual information is provided to the planner (such as explicitly labelling doorways or passages).

Planning a path through a semi-unstructured environment entails finding a physically feasible sequence of footsteps combined with whole body motions that are constrained to those footsteps which move the robot from the start state to the goal state. To apply the RPG to the problem of walking in semi-unstructured environments, we must first define the possibility exploration space for route exploration, as well as the sufficient and the necessary conditions that pertain to the robot's ability to take steps and move its whole body.

\subsection{Possibility Exploration Space}
\begin{figure}
  \centering
  \includegraphics[width=\linewidth]{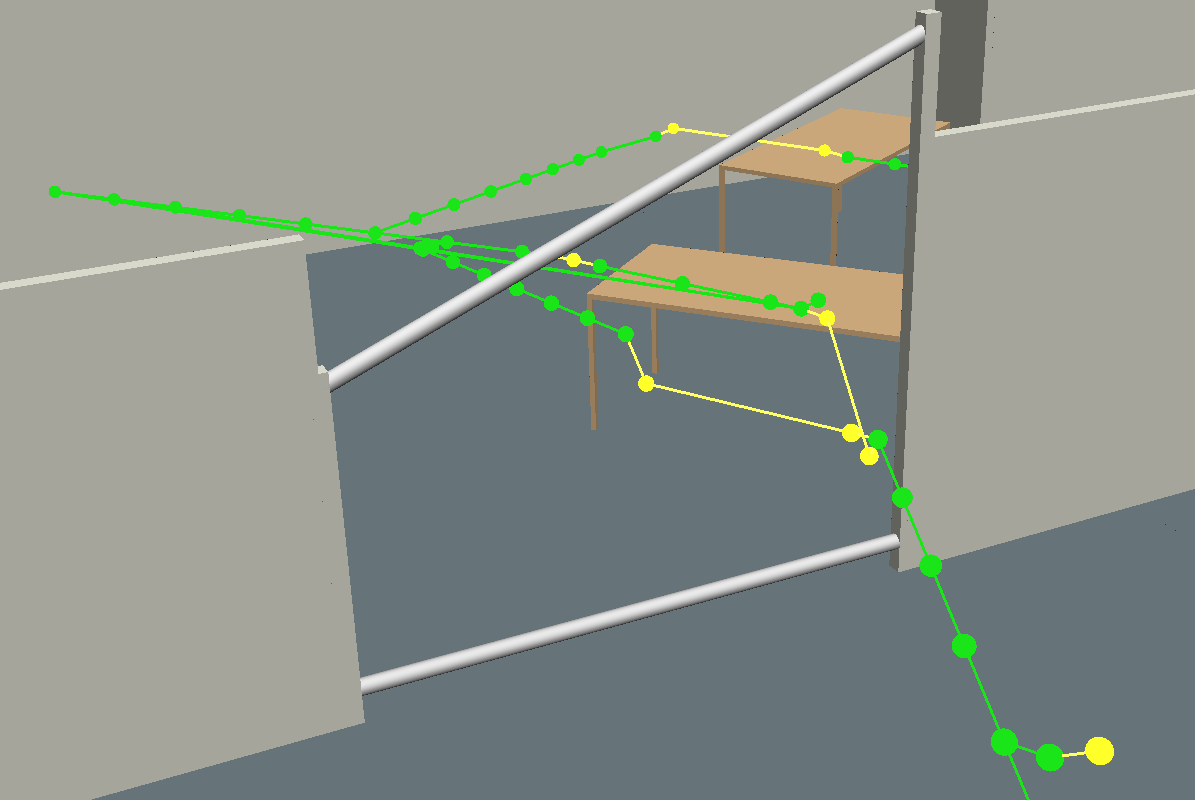}
  
  \caption{\label{fig:posgraph_snapshot}Snapshot of an RPG. Green edges are ``possible''; yellow edges are ``indeterminate''. Portions of the graph that are in wide open hallways are Possible, whereas elements that squeeze between obstacles are Indeterminate.}
\end{figure}

The vertices of the RPG exist in the possibility exploration space, $\mathscr{E}$. For the problem of walking in semi-unstructured environments, we define $\mathscr{E}$ as SE(3), which offers enough parameters to design effective sufficient and necessary conditions. Each point in $\mathscr{E}$ represents a transformation of the robot's root (pelvis) link. As we generate edges in $\mathscr{E}$, we are creating routes to guide the root link towards its destination. Figure \ref{fig:posgraph_snapshot} illustrates what a section of an RPG may look like.

\subsection{Sufficient Conditions}\label{sec:sufficient} We define the sufficient conditions by first constructing a bounding box which encapsulates all the motions that the robot might exhibit while walking or turning in any direction using some basic gait generator. The bounding geometry can be seen in Fig. \ref{fig:bounding_geometry}. If nothing in the environment is colliding with this geometry, then the robot is guaranteed to not collide with anything while performing its normal gait. For this condition, validating an edge $e(v_a,v_b)$ in SE(3) involves applying a dense sampling of SE(3) transformations from $v_a$ to $v_b$ to the bounding geometry and checking for collisions with the environment. In addition, we require the corners of the support polygon of each foot to be supported by solid ground while the robot stands in the nominal configuration shown in Fig. \ref{fig:sweep}. We can then employ a simple gait generator that is guaranteed to find a solution when these sufficient conditions are satisfied, providing us with $\pi_S$.

\subsection{Necessary Conditions} The necessary conditions are much less restrictive than the sufficient conditions. The collision geometry used for the necessary conditions can be seen in Fig. \ref{fig:minimal_geometry}. This minimal collision geometry is a subset of the actual robot's collision geometry. This subset is chosen such that it is completely unaffected by any joint motion (assuming the root link remains fixed in place). Therefore, if this geometry is in collision with any obstacles, then the robot is guaranteed to be in collision, no matter what its joint positions are. If this minimal geometry cannot sweep along an edge, then that edge is considered ``impossible'' and is left out of the RPG. Secondly, there must be at least one foothold within reach of the root link. These conditions are similar to the necessary conditions used in \cite{tonneau2015reachability}, although we leave the minimal collision geometry as it is instead of inflating it.

\subsection{Guided Multi-modal Planning}
Once the sufficient and necessary conditions are defined, the only remaining decision in our algorithm is the choice of whole body motion planner for $\pi_N$. In this work, we use Random-MMP by Hauser et. al. \cite{hauser2011randomized} because it solves motion planning problems with frequent changes in contacts. 

Random-MMP by itself can solve the semi-unstructured problems posed in this paper with the theoretical property of probabilistic completeness. However, Random-MMP performs best when it is able to utilize informed sampling. For example, when used for manipulation, Random-MMP should be provided with information on how to sample a state that is within the reachable region of the manipulatable target. In this paper, we aim to solve problems wihtout providing extra contextual information, meaning Random-MMP on its own could take prohibitively long to find a solution.

\begin{figure}
  \centering
  \begin{subfigure}[b]{\linewidth}
    \captionsetup{justification=justified}
    \centering
    \includegraphics[width=0.80\linewidth]{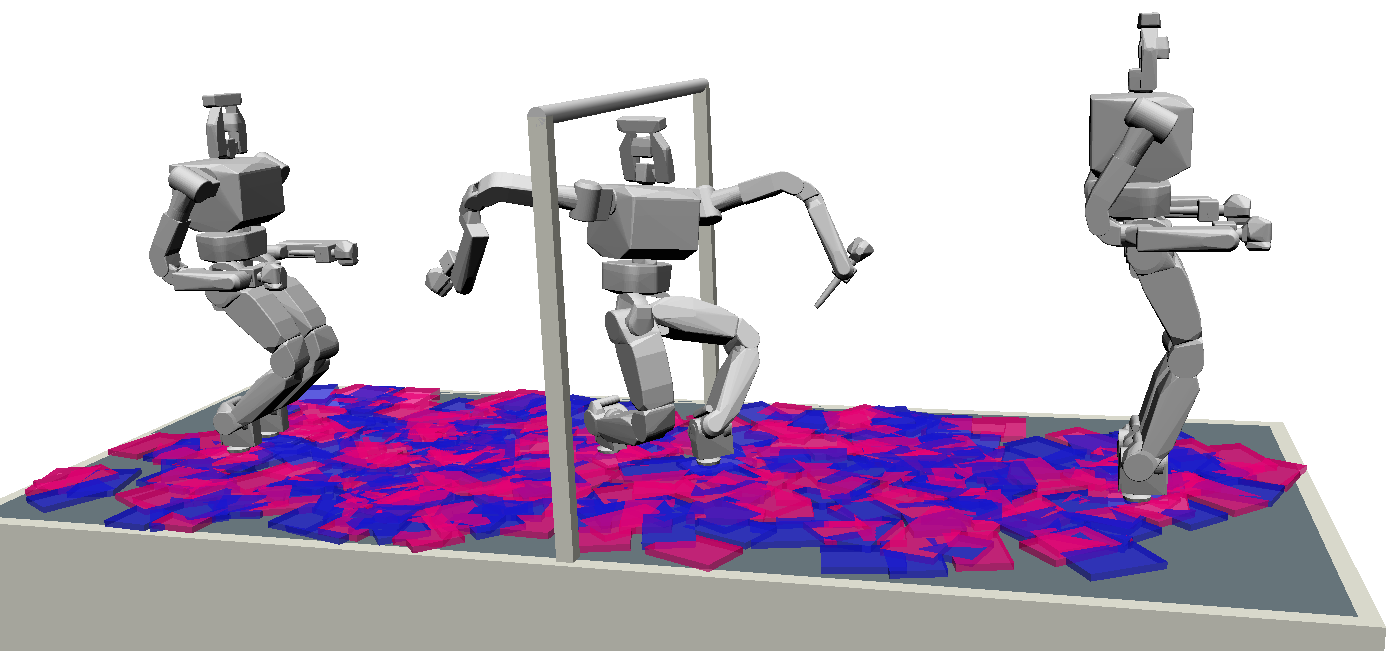}
    \caption{Uninformed Random-MMP: Exhaustive search}
    \label{fig:rmmp}
  \end{subfigure}
  
  \begin{subfigure}[b]{\linewidth}
    \captionsetup{justification=justified}
    \centering
    \includegraphics[width=0.80\linewidth]{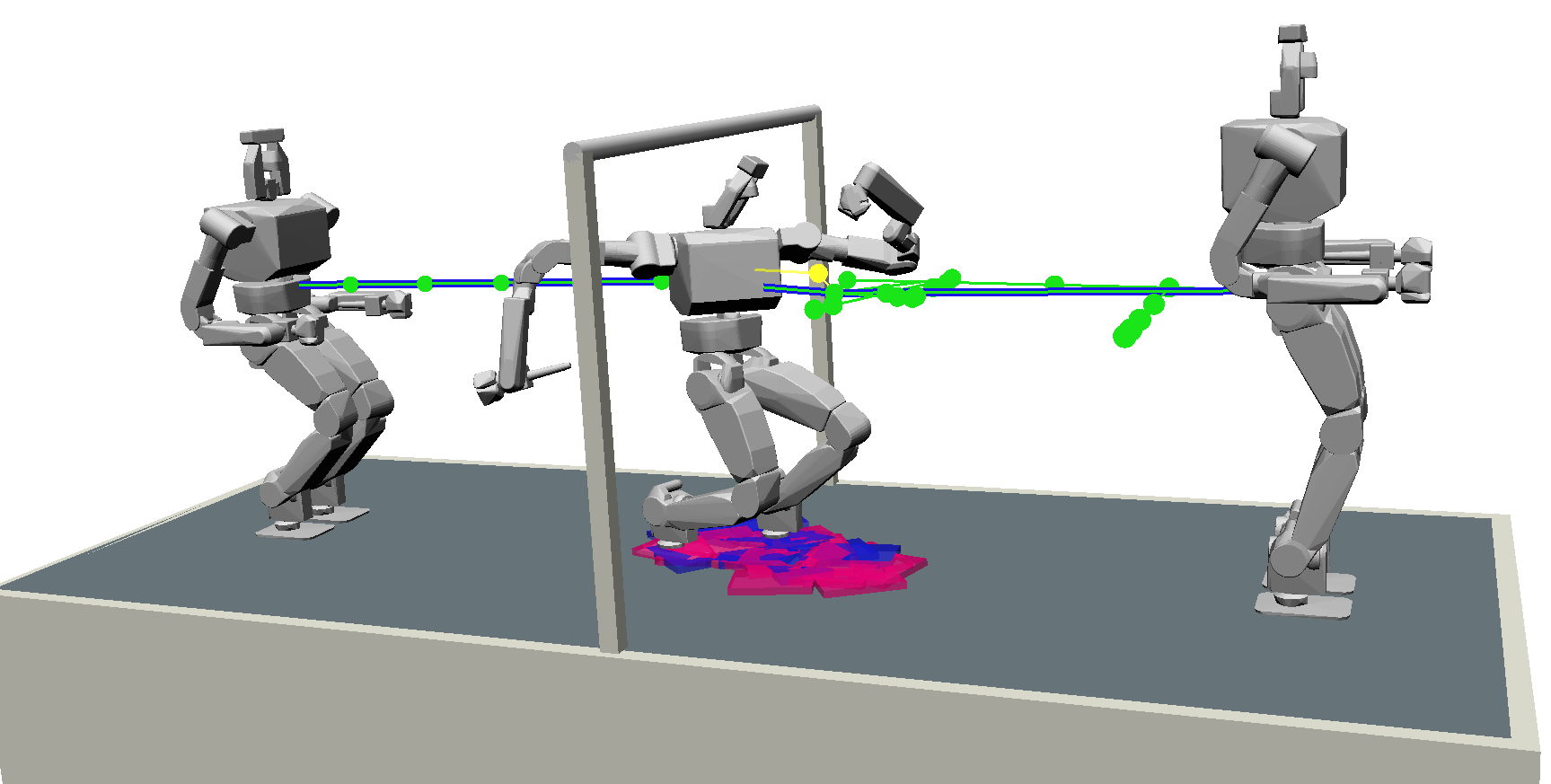}
    \caption{Guided-MMP: Focuses the search around the indeterminate (yellow) edge of the RPG.}
    \label{fig:gmmp}
  \end{subfigure}
  
  \caption{Limbo scenario. The difference in search behaviors for uninformed (a) versus guided (b) versions of Random-MMP. Blue and magenta rectangles are the foot placements of the left and right (respectively) feet for each sampled mode.}
  \label{fig:mmp}
\end{figure}

The key advantage of the RPG is that indeterminate edges can be used to guide low-level planners. Therefore, even though the original problem does not provide us with information about passages in the environment, indeterminate edges can be seen as potential passages that are worthy of further inspection. This allows us to turn an uninformed search into an informed search. We will refer to this as Guided-MMP to distinguish it from uninformed Random-MMP. Figure \ref{fig:mmp} illustrates the difference in behavior between these methods. The uninformed search takes tens of minutes whereas the guided search takes tens of seconds. The only difference between the uninformed vs. guided approaches is that the guided approach focuses its randomized mode and configuration samples to be near the indeterminate edges, according to a Normal Distribution. Meanwhile, the uninformed search broadly samples the entire domain.

\begin{table*}
\caption{\label{tab:results}Time performance results tested on an Intel\textsuperscript{\textregistered} Xeon\textsuperscript{\textregistered} Processor E3-1290 v2 (8M Cache, 3.70 GHz) with 16GB of RAM. This table shows the time spent (in seconds) and the rate of success for 30 trials of each scenario and each variation. The standard deviation is given in parentheses. Numerous trials are run because randomized planners have non-deterministic run times and non-deterministic success rates. All trials were given a one-hour timeout, at which point we consider the trial to have failed.}
\centering
\begin{tabular}{|l||l|l||l|l|}
\hline
                  & \multicolumn{2}{l||}{\textbf{Uninformed Random-MMP}} & \multicolumn{2}{l|}{\textbf{Randomzied Possibility Graph}}\\\hline
\textbf{Scenario} & \textbf{Time} & \textbf{Success} & \textbf{Time} & \textbf{Success}\\\hline\hline
Limbo Scenario (Fig. \ref{fig:mmp})             & 2355.5 (1000.3) & 86.67\%  & 20.56 (26.51)    & 100\%\\\hline\hline
Four Routes Scenario (Fig. \ref{fig:layout})    & 3600 (0)        & 0\%      & 68.44 (33.98)    & 96.67\%\\\hline
West Door Blocked (Fig. \ref{fig:west_blocked}) & 3600 (0)        & 0\%      & 295.54 (205.847) & 96.67\%\\\hline
East Door Blocked (Fig. \ref{fig:east_blocked}) & 3600 (0)        & 0\%      & 349.60 (195.38)  & 100\%\\\hline
Bars Removed (Fig. \ref{fig:no_bars})           & 3600 (0)        & 0\%      & 7.31 (5.73)      & 100\%\\\hline
\end{tabular}
\end{table*}

\section{Experiments}\label{sec:experiments}

We present two virtual experimental scenarios. The first scenario, shown in Fig. \ref{fig:mmp}, is referred to as the ``Limbo Scenario''. The robot must traverse from one side of the platform to the other while ducking underneath a limbo bar. It demonstrates the performance difference between an uninformed Random-MMP search versus Guided-MMP which is supplemented by the Randomized Possibility Graph. Even in such a simple scenario, the RPG improves performance by two orders of magnitude.

The second scenario is referred to as the ``Four Routes Scenario''. It demonstrates some of the larger scale capabilities of the RPG. The robot must navigate from the southwest corner to the northeast corner of the floor. Each room has two entrances, making a total of four possible routes the robot may choose from. Each doorway has its own challenges associated with passing through it, seen in Fig. \ref{fig:doorways}.

\begin{figure}
  \centering
  \begin{subfigure}[b]{\linewidth}
    \captionsetup{justification=centering}
    \centering
    \includegraphics[width=0.63\linewidth]{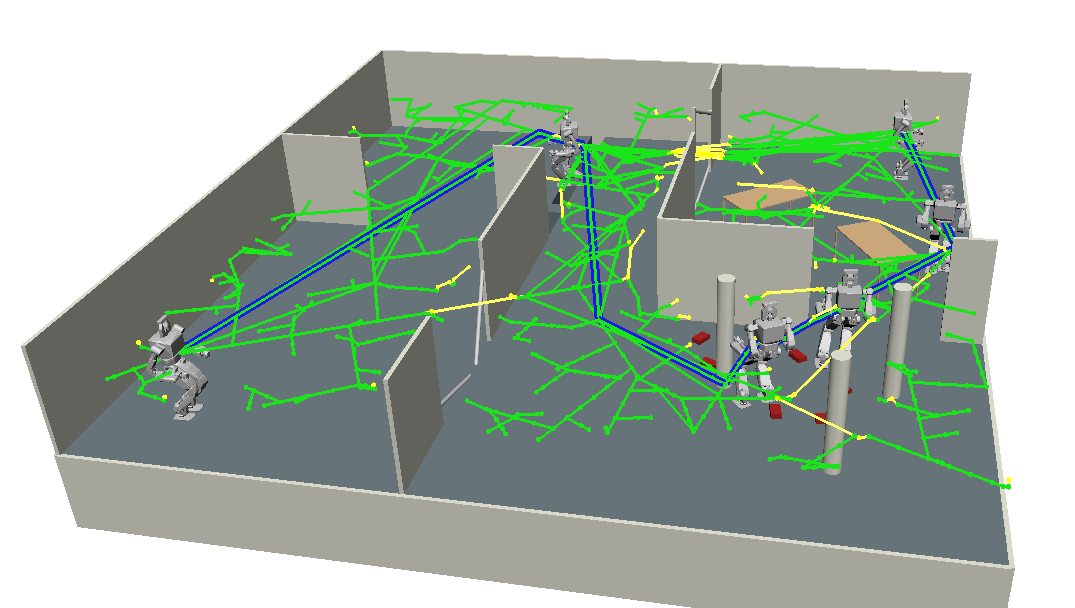}
    \caption{Solution for the standard variation of the scenario}
    \label{fig:all_open}
  \end{subfigure}
  \begin{subfigure}[b]{\linewidth}
    \captionsetup{justification=centering}
    \centering
    \includegraphics[width=0.63\linewidth]{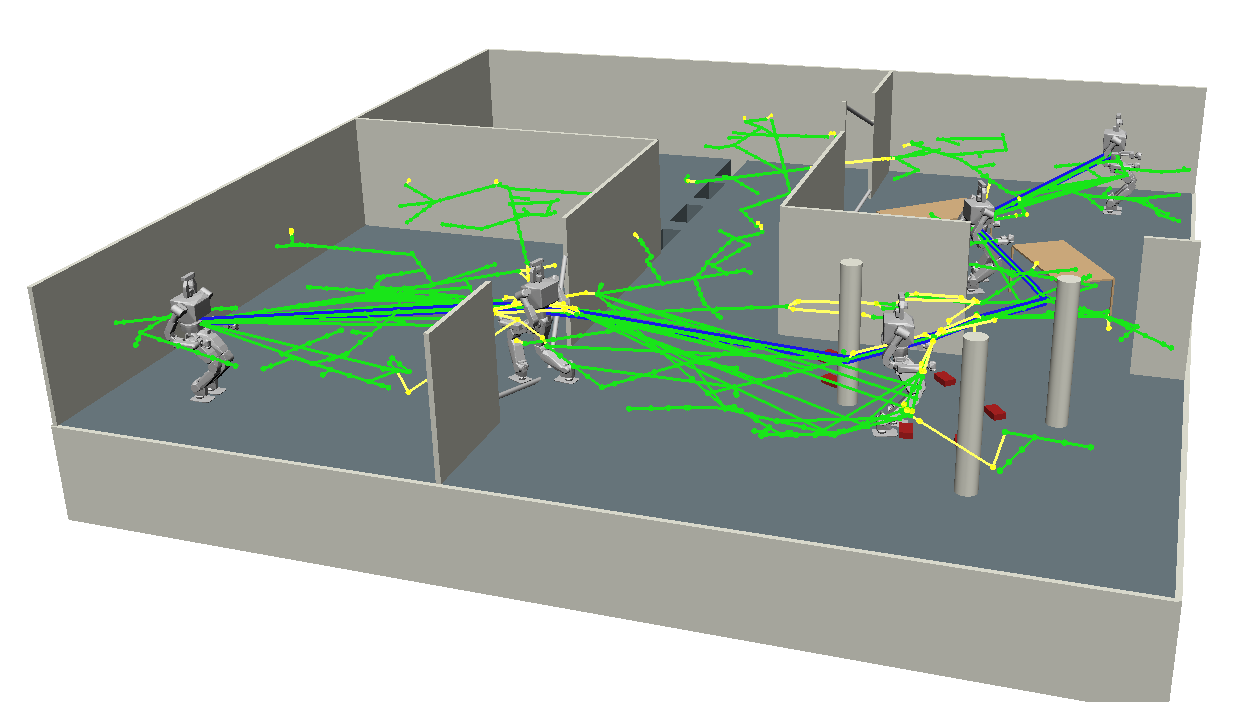}
    \caption{Solution when the west doorway is blocked off}
    \label{fig:west_blocked}
  \end{subfigure}
  
  \begin{subfigure}[b]{\linewidth}
    \captionsetup{justification=centering}
    \centering
    \includegraphics[width=0.63\linewidth]{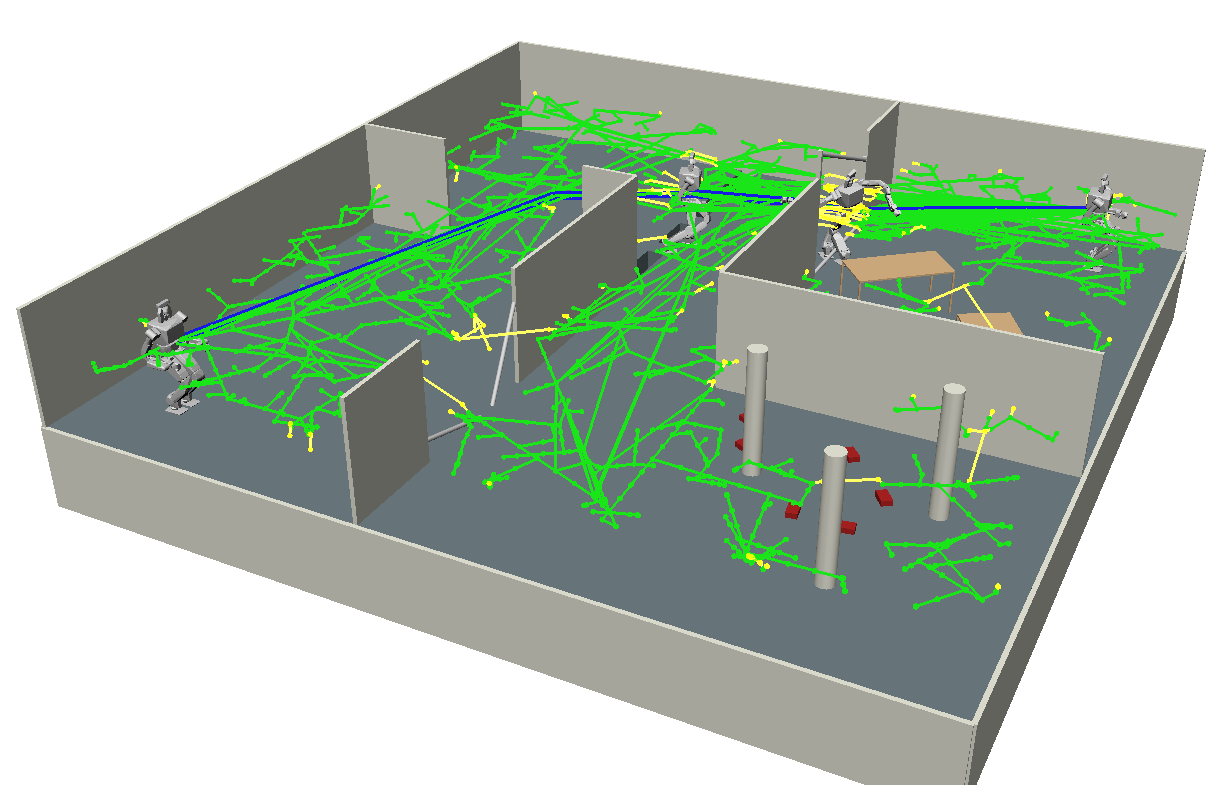}
    \caption{Solution when the east doorway is blocked off}
    \label{fig:east_blocked}
  \end{subfigure}
  \begin{subfigure}[b]{\linewidth}
    \captionsetup{justification=centering}
    \centering
    \includegraphics[width=0.63\linewidth]{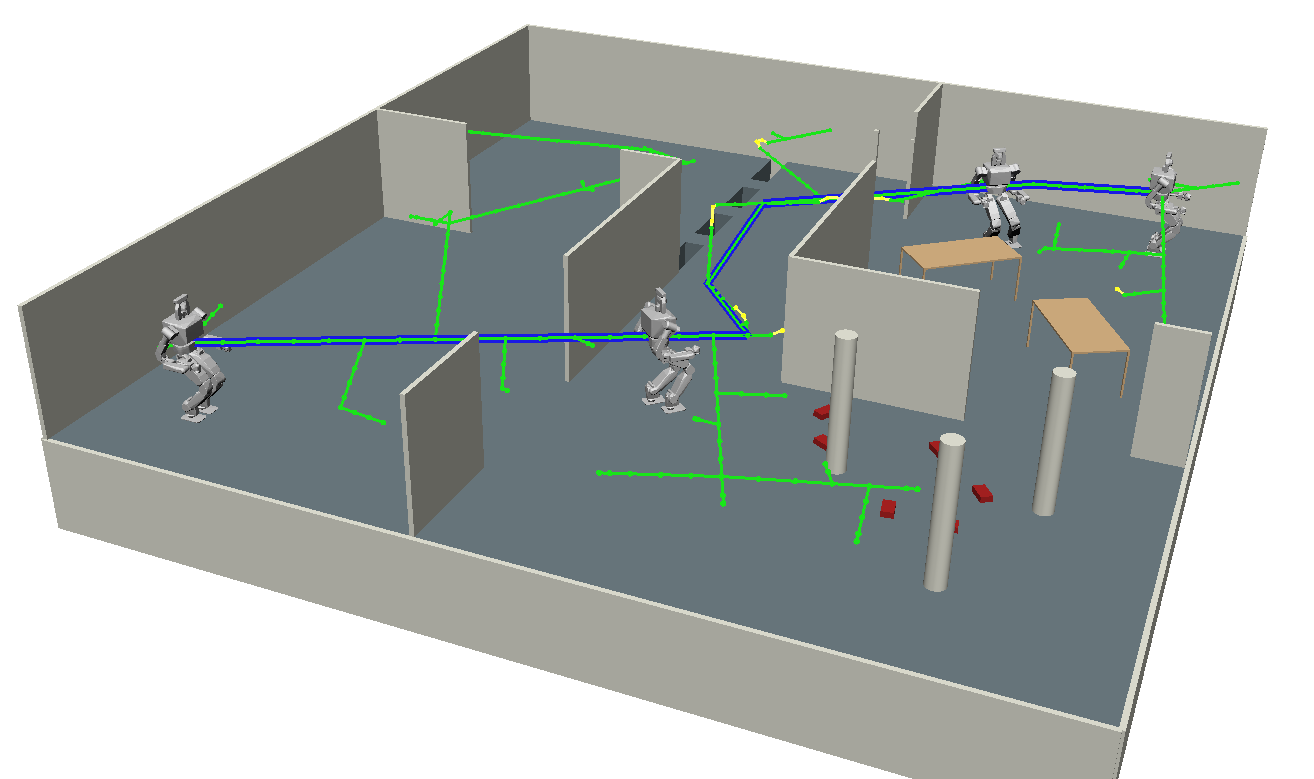}
    \caption{Solution with bars removed from the north and south}
    \label{fig:no_bars}
  \end{subfigure}
  
  \caption{Variations of the ``Four Routes Scenario''.}
  \label{fig:alternatives}
\end{figure}

The route which is easiest for the planner to find passes through the west doorway, down the hallway, and then through the east doorway as shown in Fig \ref{fig:all_open}. Those doorways are the easiest to plan for because they have the most ``expansiveness'' as defined in \cite{hsu1997path}. Since the planner chooses the first successful plan that it finds, this will be the most commonly chosen route (although the randomized nature of the planner does not guarantee that this will always be the chosen route). We can tweak the environment to force the robot to find a path through the south doorway by blocking off the west doorway (as in Fig. \ref{fig:west_blocked}). Similarly, the robot can be forced to find a path through the north doorway by blocking off the east (as in Fig. \ref{fig:east_blocked}). Each of these modifications result in considerably longer run times as shown in Table \ref{tab:results}. This shows that when the planner is not required to find routes through challenging regions, it will naturally tend to favor regions where it can find paths easily.

In some cases, there may be a clear path from the start to the goal. To simulate this, in Fig. \ref{fig:no_bars} we remove the bars that are obstructing the north and south doorways. The result is a lean graph and a short plan time as seen in Table \ref{tab:results}.


\section{Conclusions}\label{sec:conclusion}

\begin{figure}
  \centering
  \includegraphics[width=0.84\linewidth]{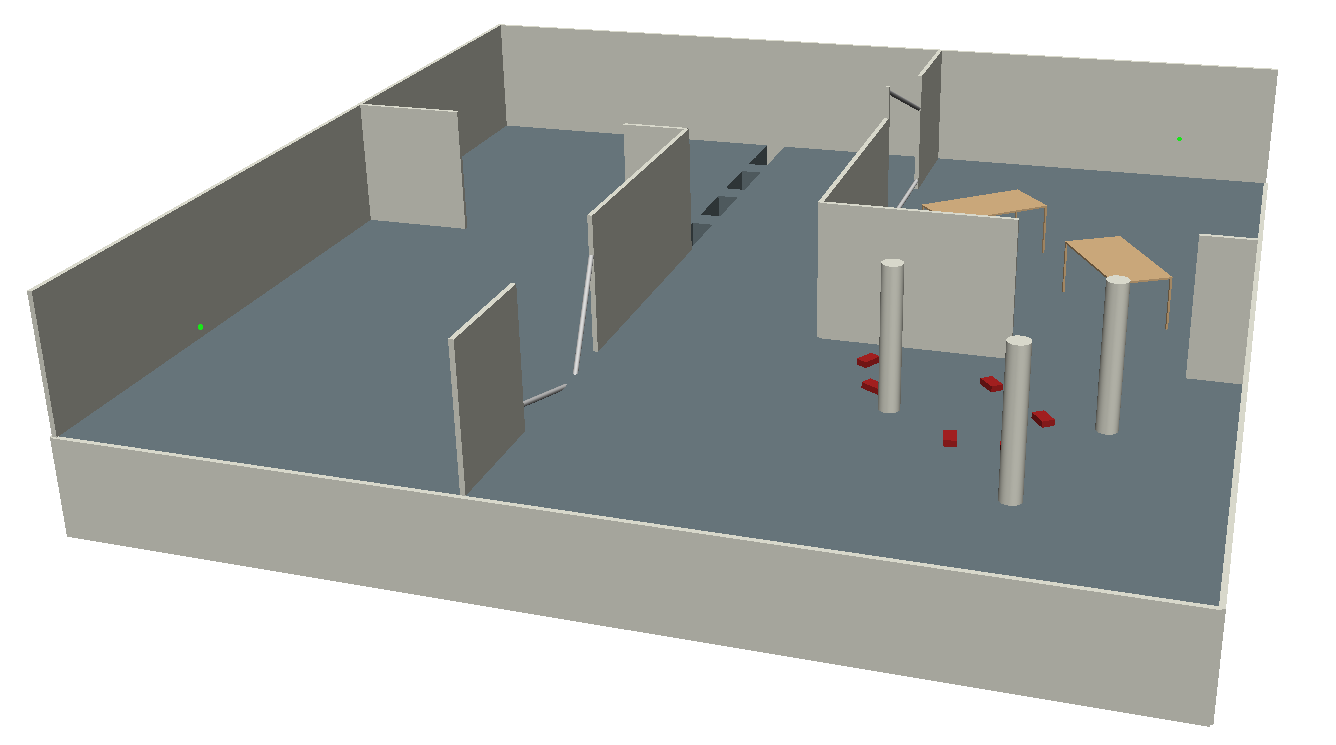}
  \caption{Basic ``Four Routes'' scenario.}
  \label{fig:layout}
\end{figure}

This work presented a new high-level algorithm for identifying routes in semi-unstructured environments, and a way to leverage this information in lower-level motion planners. The method presented improves performance of existing algorithms by orders of magnitude, making large-scale problems tractable when they would not have been previously.

The use of necessary conditions and guide routes shares some similarities to the reachability-based planner of Tonneau, et. al. \cite{tonneau2015reachability}. The key novelty of this work is that we also leverage sufficient conditions, which are not used by Tonneau. Furthermore, we use probabilistically complete whole body planners when evaluating the indeterminate portions of the guide routes, which we believe allows the robot to more fully utilize its maneuverability, at the cost of longer run times. In future work, we will apply the RPG to uneven terrain, at which point a direct comparison between the RPG and the reachability-based planner will be possible.

Future work will also investigate other ways that the RPG can be leveraged to generate plans. For example, the graph could consider the possibilities of dynamic actions rather than just quasi-static actions. To improve the evaluation of indeterminate edges, it would be natural to use a variety of low-level planners in parallel that could compete to confirm indeterminate edges rather than using a single catch-all probabilistically complete low-level planner which might not perform particularly well in all scenarios. We may also be able to define some distinct variations of sufficient conditions where each variation corresponds to a different flavor of gait generator, allowing a broader range of edges to be covered by sufficient conditions.

The completeness of the RPG is currently being investigated. Future work will examine under what conditions the RPG may be considered probabilistically complete. This will likely depend on the completeness of the low-level planners that it can utilize.

\newpage
\section*{Acknowledgments}

This work was supported by DARPA grant D15AP00006.

\bibliographystyle{IEEEtran}
\bibliography{icra2017.bib}

\end{document}